\documentclass[10pt,twocolumn,letterpaper]{article}

\usepackage{iccv}
\usepackage{times}
\usepackage{epsfig}
\usepackage{graphicx}
\usepackage{amsmath}
\usepackage{amssymb}
\usepackage[usenames,dvipsnames]{color}
\usepackage[outline]{contour}

\usepackage{ctable}
\usepackage{bm}

\usepackage[pagebackref=true,breaklinks=true,letterpaper=true,colorlinks,bookmarks=false]{hyperref}

\iccvfinalcopy 

\def\iccvPaperID{535} 

\newcommand\ig[2]{\includegraphics[width=#1\linewidth]{./figures/#2}}
\newcommand\classname[1]{\textsl{#1}}
\newcommand\franchise[1]{\textsl{#1}}
\newcommand\beatevent{beat-event}
\newcommand\beatevents{beat-events}

\ificcvfinal\fi
\begin{document}

\title{Beat-Event Detection in Action Movie Franchises}

\author{Danila Potapov
\and
Matthijs Douze
\and
Jerome Revaud
\and
Zaid Harchaoui
\and
Cordelia Schmid
}

\maketitle

\begin{abstract}

While important advances were recently made 
towards temporally localizing and recognizing specific human actions or activities in videos, 
efficient detection and classification of long video chunks belonging to semantically-defined categories 
such as ``pursuit'' or ``romance'' remains challenging.

We introduce a new dataset, \textbf{Action Movie Franchises}, consisting of a 
collection of Hollywood action movie franchises.
We define 11 non-exclusive semantic categories~---
called \textbf{beat-categories}~--- that are broad enough to cover most of the movie footage.
The corresponding \textbf{\beatevents{}} are annotated as groups of video shots, possibly overlapping.
We propose an approach for localizing \beatevents{} based on 
classifying shots into beat-categories and learning the temporal 
constraints between shots.  We show that temporal constraints 
significantly improve the classification performance.  We set up an evaluation
protocol for beat-event localization as well as for shot classification,
depending on whether movies from the same franchise are present or not
in the training data.  

\end{abstract}


\section{Introduction}

Automatic understanding and interpretation of videos is a challenging and important
problem due to the massive increase of available video data,
and the wealth of semantic variety of video content. Realistic videos
include a wide variety of actions, activities, scene type, etc.
During the last decade, significant progress
has been made for action retrieval and recognition of specific, 
stylized, human actions. In particular, powerful visual features were proposed 
towards this goal~\cite{OVS13,oneata:hal-01074442,hengwangICCV13}. For more general 
types of events in videos, such as activities, efficient approaches 
were proposed and benchmarked as part of the TrecVid Multimedia Event Detection
(MED) competitions~\cite{2014trecvidover}.  State-of-the-art approaches combine
features from all modalities (text, visual, audio), static and motion features
(possibly learned beforehand with deep learning), and appropriate fusion
procedures. 

\newcommand{\igname}[2]{%
\includegraphics[width=2.7cm]{figures/categories/#1.jpg}%
\makebox[0pt]{\hspace*{-26mm}%
\textcolor{white}{\contour{black}{\small\textsf{#2}}}%
}%
}

\begin{figure}
\centering
  \begin{tabular}{c@{\,}c@{\,}c}
\igname{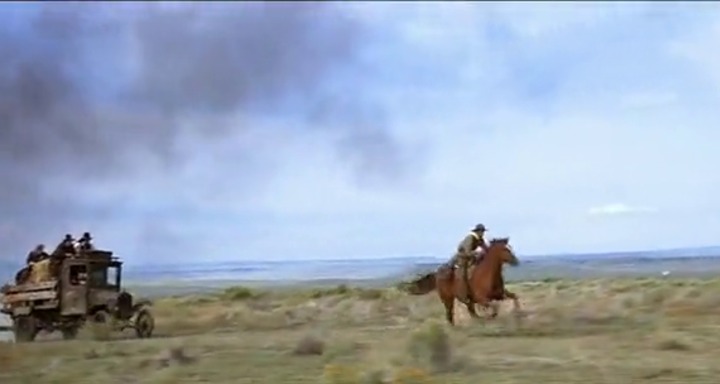}{pursuit} & 
\igname{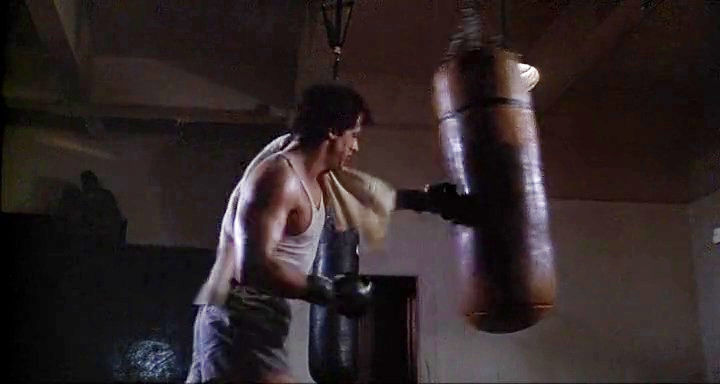}{battle preparation} & 
\igname{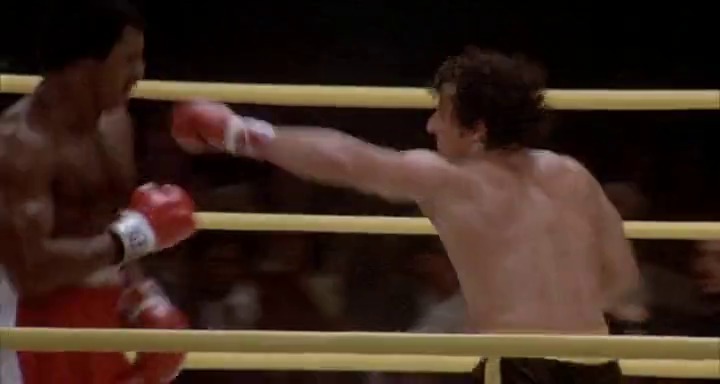}{battle} \\
\igname{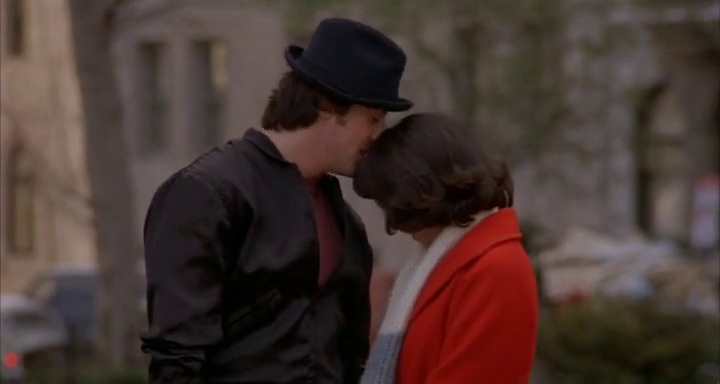}{romance} & 
\igname{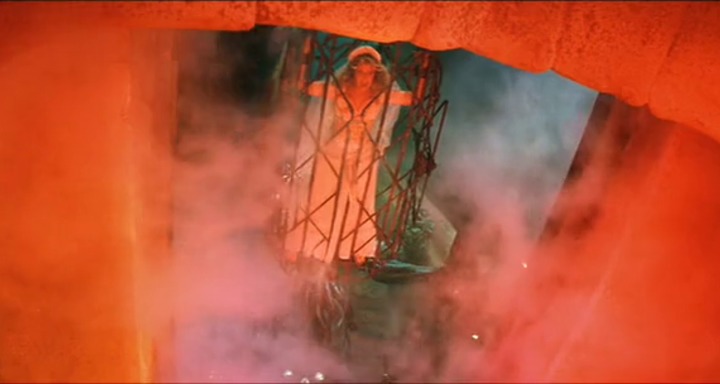}{despair good}& 
\igname{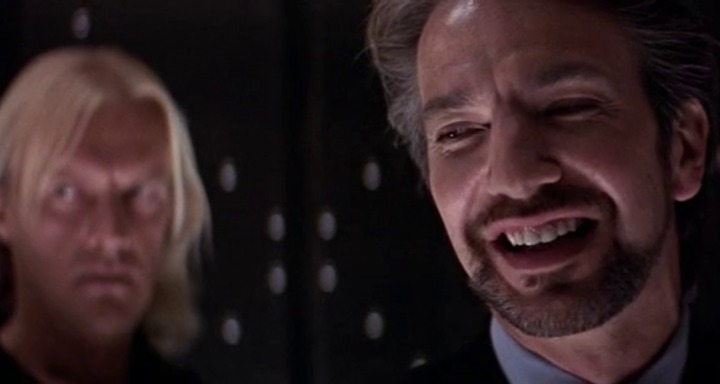}{joy bad} \\
\igname{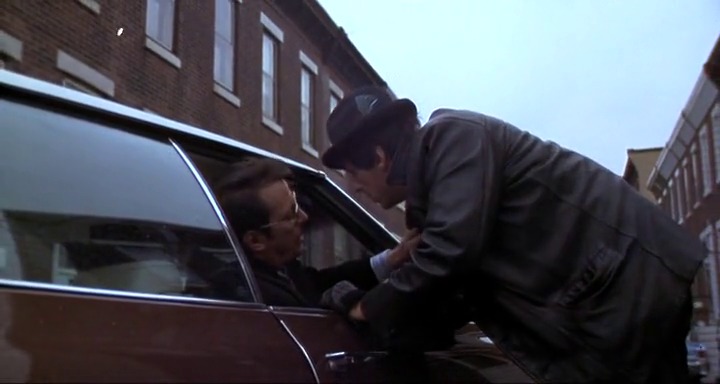}{good argue good} & 
\igname{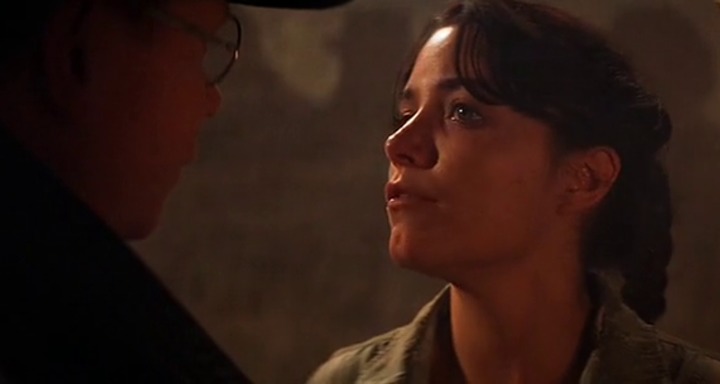}{good argue bad} & 
\igname{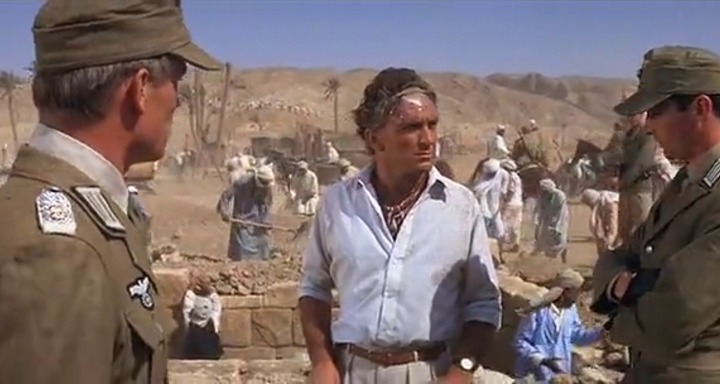}{bad argue bad} \\
\igname{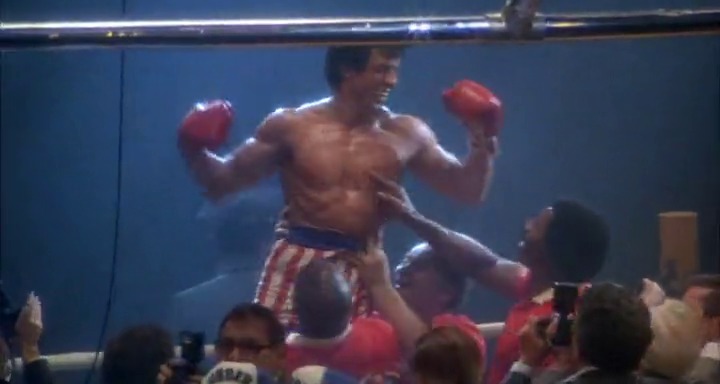}{victory good} & 
\igname{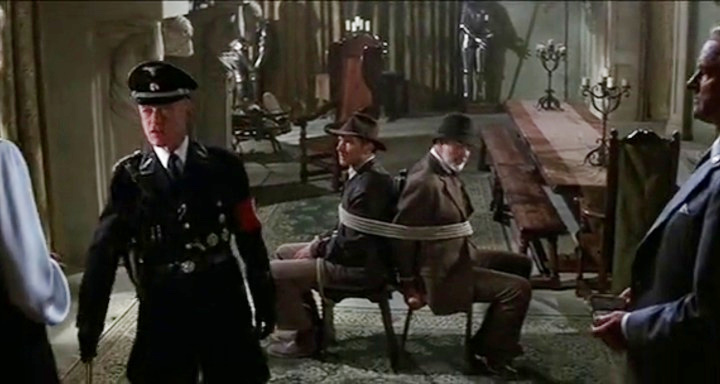}{victory bad} & 
\igname{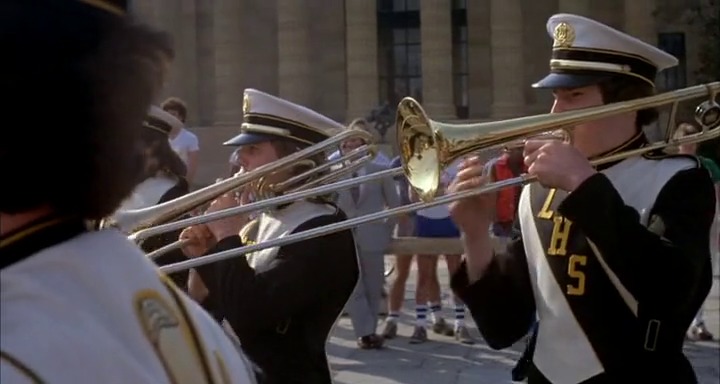}{NULL}\\
  \end{tabular} 
    \caption{Example frames for the categories from the Action Movie Franchises dataset.}
    \label{fig:categories}
\end{figure}

\begin{figure}
    \includegraphics[width=\linewidth]{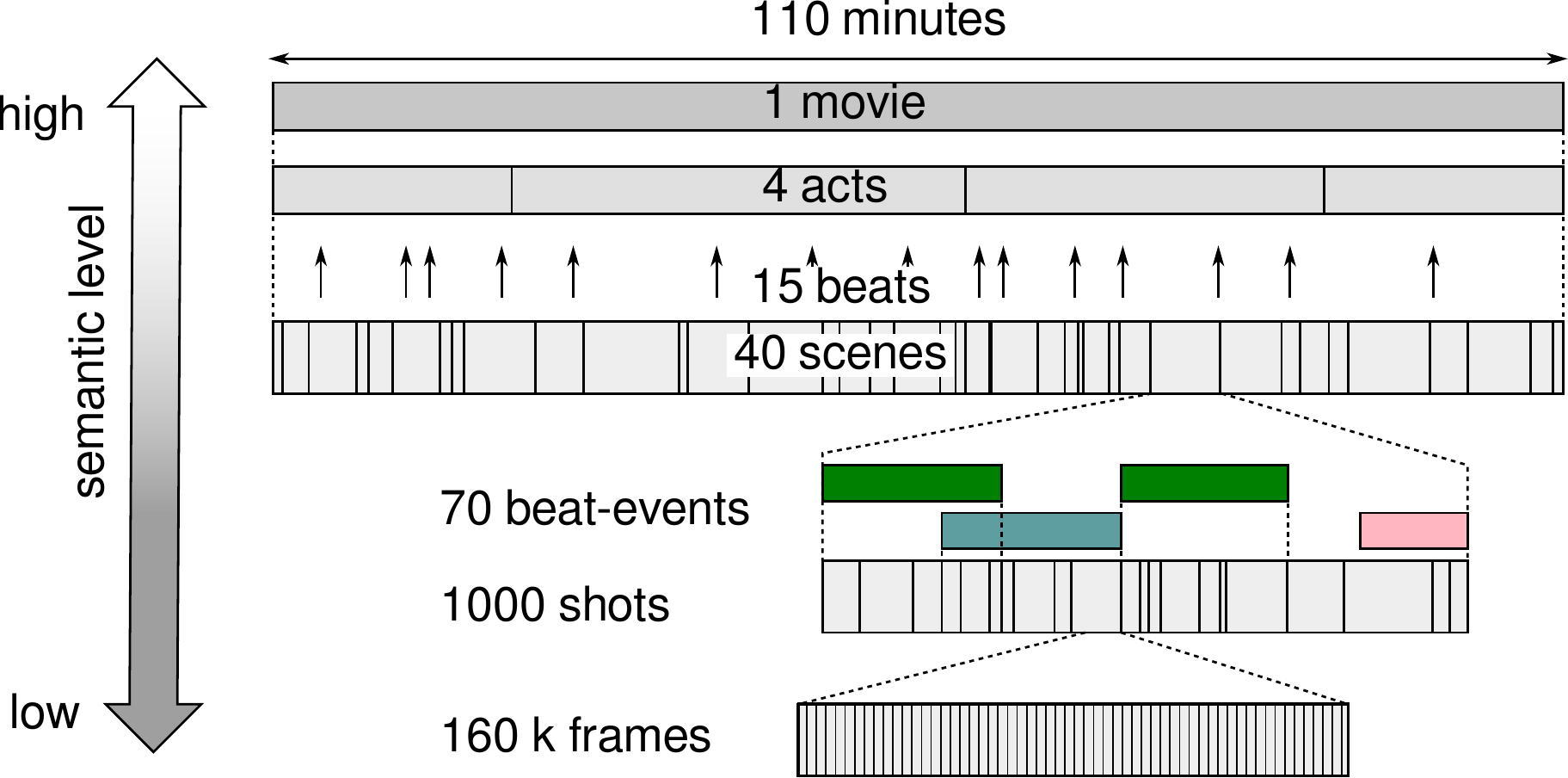}
    \caption{\label{fig:stctaxo}
      Temporal structure of a movie, according to the taxonomy of ``Save the Cat''~\cite{stc}, and our level of annotation, the \beatevent{}.
    }
\end{figure}

In this work, we aim at detecting events of the same semantic level as Trecvid MED, but on
real action movies that follow a structured scenario.
From a movie script-writer's point of view~\cite{stc,ronfard:inria-00423417},
a Hollywood movie is more or less constrained to a  
set of standard story-lines.
This standardization helps matching the audience
expectations and habits. However, movies need to be fresh and novel enough to fuel the interest
of the audience. So, some variability must be introduced in the story lines to maintain the interest.
Temporally, movies are subdivided in a hierarchy of acts, scenes,
shots, and finally, frames (see Figure~\ref{fig:stctaxo}). Punctual changes in the
storyline give it a rhythm. They  are  called ``beats'' and are common to many films. 
A typical example of beat is the moment when an unexpected solution saves the hero.

From a computer vision point of view, frames are readily available and
reliable algorithms for shot detection exist. Grouping shots into scenes is harder.
Scenes are characterized by a uniform location, set of characters or
storyline. The semantic level of beats and acts is out of reach.
We propose here to attack the problem on an intermediate level by detecting ``\beatevents{}''.
Temporally, they consist in sequences of consecutive shots  
and typically last a few minutes.  
Shots offer a suitable granularity, because movies are 
edited so that they follow the rhythm of the action. 
Semantically, they are of 
a higher level than the actions in most current benchmarks, but 
lower than the beats, which are hard to identify even for a human.

For the purpose of research, we built an annotated dataset of Hollywood action movies, called \textbf{Action Movie Franchises}. 
It comprises $20$ action movies from 5 franchises: \franchise{Rambo}, \franchise{Rocky}, \franchise{Die Hard},  
\franchise{Lethal Weapon}, \franchise{Indiana Jones}. A movie franchise 
refers to a series of movies on the same ``topic'', sharing similar story lines
and the same characters. 
In each movie, we annotate shots into several non-exclusive beat-categories.
We then create a higher level of annotation, called beat-events, which consists
of consistent sequences of shots labeled with the same beat-category.

Figure~\ref{fig:categories} illustrates the beat-categories that we use in our dataset. 
They are targeted at action movies and, thus, rely on semantic categories that often reply 
on the role of the characters, such as hero (good) or villain (bad). 
We now briefly describe all categories.
First, we define three different action-related beat-categories:
\emph{pursuit}, \emph{battle preparation} and \emph{battle}, shown in the first row of Fig.~\ref{fig:categories}.
We also define categories centered on the emotional state of the main characters:
\emph{romance},  
\emph{despair good} (\eg when the hero thinks that all is lost) and 
\emph{joy bad} (\eg when the villain thinks he won the game), 
see second row of Fig.~\ref{fig:categories}.
We also include different categories of dialog between all combinations of good and bad characters: 
\emph{good argue good}, \emph{good argue bad} and
\emph{bad argue bad} (third row of Fig.~\ref{fig:categories}).
Finally, we add two more categories notifying a temporary victory
of a good or bad character (\emph{victory good} and \emph{victory bad}, last row of Fig.~\ref{fig:categories}).
We also consider a NULL category, corresponding to shots that can not 
be classified into any of the aforementioned beat-categories.

\let\olditemize\itemize
\renewcommand{\itemize}{
  \olditemize
  \setlength{\itemsep}{1pt}
  \setlength{\parskip}{0pt}
  \setlength{\parsep}{0pt}
  \settowidth{\labelwidth}{{\small$\bullet$}}
 \setlength{\labelsep}{0.5em}%
}

In summary, we introduce the \textbf{Action Movie Franchises} dataset, which features dense annotations of 11 beat-categories in 20 action movies
at both shot and event levels.  
To the best of our knowledge, a comparable dense annotation of videos does not exist. 
 
The semantic level of our beat-categories will drive progress in action recognition towards new approaches 
based on human identity, pose, interaction and semantic audio features. State-of-the-art methods are without doubt not sufficient for such categories. 
Action movies and related professionally produced content account for a major 
fraction of what people watch on a daily basis. 
There exists a large potential for applications, such as access to video archives and movie 
databases, interactive television and automatic annotation for the shortsighted.

Furthermore, we define several evaluation protocols, to investigate the impact of
franchise-information (testing with or without previously seen movies
from the same franchise) and the performance for both classification and localization tasks.
We also propose an approach for classification of video shots into beat-categories based on 
a state-of-the-art pipeline for multimodal feature extraction, classification and fusion. Our approach for localizing beat-events uses 
a temporal structured inferred by a conditional random field (CRF) model learned from training data.

We will make the Action Movie Franchises dataset publicly available upon publication to the research community to further advance video understanding.


\section{Related work}

\begin{table*}
\resizebox{\linewidth}{!}{
\begin{tabular}{|l|cc|cr|rrrr|}
\hline
Name & \# classes &  example & annotation & \# train & \multicolumn{4}{c|}{durations}  \\
     &            &  class    &  unit    &   units       & avg unit & annot & NULL & coverage \\
\hline
\hline
\multicolumn{9}{|c|}{Classification}\\
\hline
UCF 101~\cite{soomro2012ucf101} & 101 & high jump  & clip & 13320 & 7.21s & 26h39 & 0h & - \\ 
HMDB 51~\cite{kuehne2011hmdb} & 51 &brush hair  & clip & 6763 & 3.7s &6h59 & 0h & - \\ 
TrecVid MED 11 & 15 & birthday party &   clip & 2650 & 2m54 &128h&  315h &  29\% \\ 
Action Movie Franchises & 11 & good argue bad &  shot & 16864 & 5.4s & 25h29 & 15h42 & 57.1\% \\ 
\hline
\hline
\multicolumn{9}{|c|}{Localization}\\
\hline
Coffee \& Cigarettes & 2 & drinking &  time interval & 191 & 2.2s &7m12s & 3h26 & 3.3\% \\ 
THUMOS detection 2014 & 20 & floor gymnastics  & t.i. on clip & 3213 & 26.2s & 3h22 & 167h54 & 2.0\% \\ 
MediaEval VSD~\cite{demartyVSD} & 10 & fighting & shot/segment & 3206 & 3.0s & 2h38 & 55h20 & 4.5\% \\ 
Action Movie Franchises & 11 & good argue bad &  \beatevent{} & 2906 & 35.7s & 28h49 & 14h08 & 61.4\% \\ %
\hline
\end{tabular}
}
\caption{Comparison of classification and localization datasets. 
annot = total duration of all annotated parts; 
NULL = duration of the non-annotated (NULL or background) footage; 
coverage = proportion of annotated video footage. 
\label{tab:actionmoviesstats}}
\end{table*}

\paragraph{Related datasets.}
Table~\ref{tab:actionmoviesstats} summarizes recent state-of-the-art
datasets for action or activity recognition.  
Our Action Movie Franchises dataset mainly differs from existing ones with
respect to the event complexity  and the density  of annotations. 
Similarly to Coffee \& Cigarettes and MediaEval Violent Scene
Detection (VSD), our Action Movie Franchises dataset is built on professional
movie footage. However, while the former datasets 
only target short and sparsely occurring events, we provide dense annotations 
of  \beatevents{} spanning larger time intervals. 
Our beat-categories are also of significantly higher semantic level than
those in action recognition datasets like Coffee \& Cigarettes, 
UCF~\cite{soomro2012ucf101} and HMDB~\cite{kuehne2011hmdb}. 
A consequence is that our dataset remains very challenging
for state-of-the-art algorithms, as shown later in the experiments. 
Events of a similar complexity can be found in TrecVid MED 2011--2014~\cite{2014trecvidover}, but
our dataset includes precise temporally localized annotations.

\vspace{-0.3cm}
\paragraph{Action detection in movies.}
Action detection (or action localization), that is finding if and when a
particular type of action was performed in long and unsegmented video streams,
received a lot of attention in the last decade. The problem was considered in a
variety of settings: from still images~\cite{raptis2013poselet}, from
videos~\cite{gaidon:hal-00804627,hengwangICCV13}, with or without weak
supervision, etc. Most works focused on highly stylized human actions such as
``open door'', ``sit down'', which are typically \emph{temporally salient} in
the video stream.

Action or activity recognition can often be boosted using temporal reasoning on
the sequence of atomic events that characterize the action, as well as the
surrounding events that are likely to precede or follow the action/activity of
interest. We shall only review here the
``temporal context'' information from surrounding events; the decomposition of
action or activities into sequence of atomic events~\cite{gaidon:hal-00804627}
is beyond the scope of our paper. Early works along this
line~\cite{rui1998exploring} proposed to group shots and organize groups into
``semantic'' scenes, each group belonging exclusively to only one scene. Results
were evaluated subjectively and no user study was conducted.

Several papers proposed to use movie (or TV series) scripts to leverage the
temporal structure~\cite{everingham2006buffy,marszalek2009actions}.
In~\cite{marszalek2009actions}, movie scripts are used to obtain scene and
action annotations. Retrieving and exploiting movie scripts can be tricky and
time-consuming. In many cases, movie scripts are simply not
available.  Thus, we
did not use movie scripts to build our dataset and do not consider this
information for training and testing. However, we do use another modality, the
audio track, in a systematic way, and perform fusion following state-of-the-art
approaches in multimedia~\cite{li2004content}, and TrecVid
competitions~\cite{2014trecvidover}.

In~\cite{cour2008alignment}, the authors structure a movie into a sequence of
scenes, where each scene is organized into interlaced threads. An efficient
dynamic programming algorithm for structure parsing is proposed. Experimental
results on a dataset composed of TV series and a feature-length movie are
provided.  More recently, in~\cite{bojanowski2013movies}, actors
and their actions are detected simultaneously under weak supervision of movies
scripts using discriminative clustering. Experimental results on 2 movies
(\franchise{Casablanca}  and  \franchise{American beauty}) are presented, for 3
actions (\classname{walking}, \classname{open door} and \classname{sit down}).
The approach improves person naming compared to previous methods. In this work,
we do not use supervision from movie scripts to learn and uncover the
temporal structure, but rather learn it directly using a conditional random
field that takes SVM scores as input features.  The proposed
approach is more akin to~\cite{hoai2011joint}, where joint segmentation and
classification of human actions in video is performed on toy
datasets~\cite{hoai2012max}.


\vspace{-0.1cm}
\section{Action Movie Franchises}  

We first describe the \emph{Action Movie Franchises} dataset and the annotation
protocol.  Then, we highlight some striking features in the structure of the
movies observed during and after the annotation process. Finally, we propose an
evaluation protocol for shot classification into beat-categories and for \beatevent{} localization. 

\begin{figure*}
\centering
\ig{1.0}{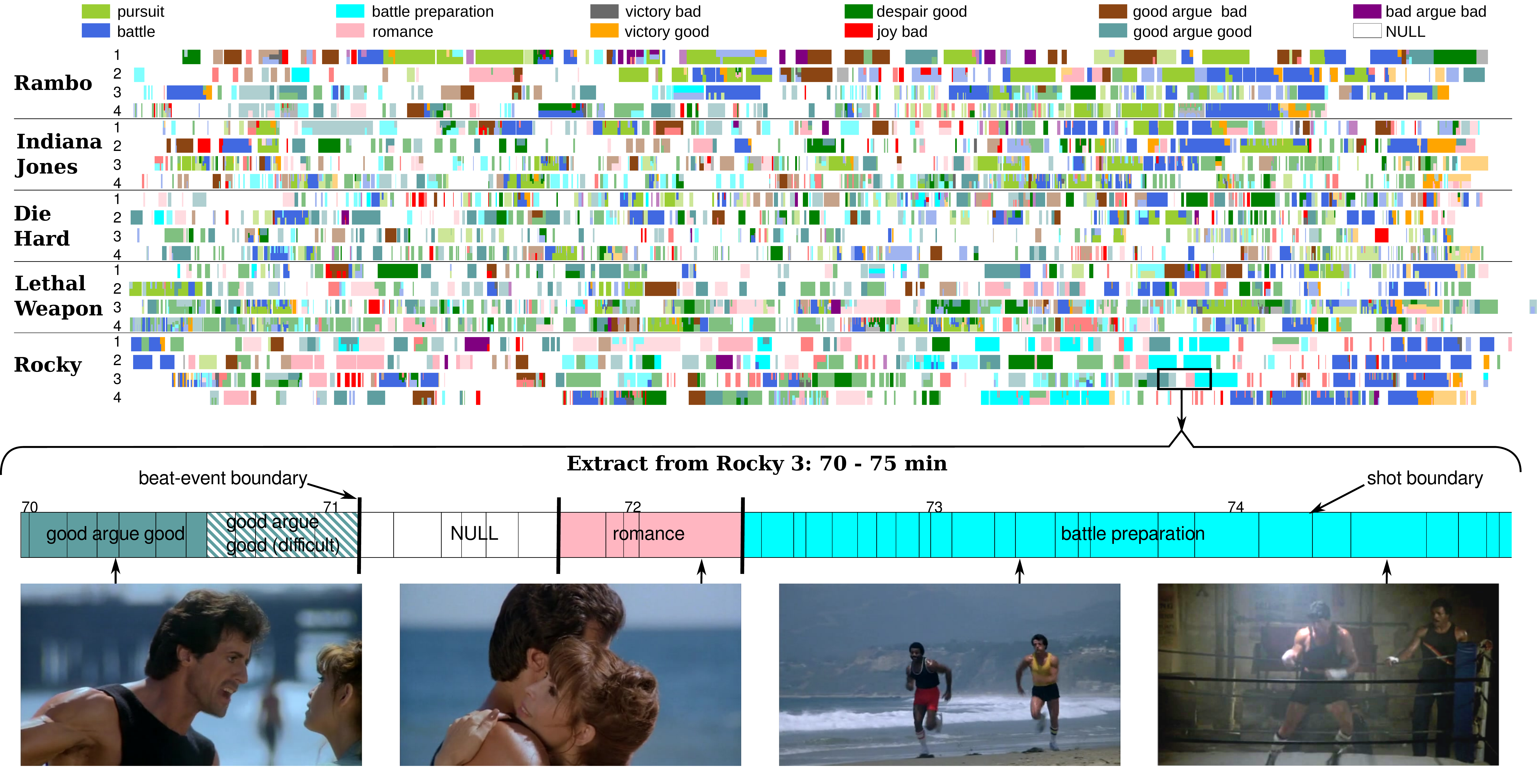}

\caption{Top: Beat-events annotated for the Action Movie Franchises dataset, one movie per line, plotted along the temporal axis. All the movies were scaled to the same length.  Bottom: zoom on a movie extract showing the shot segmentation, the annotations and the beat-events. Best viewed onscreen.}
\label{fig:instances}
\end{figure*}

\subsection{The movies}

The Action Movie Franchises dataset consists of 20 Hollywood action movies
belonging to 5 famous franchises:
\franchise{Rambo, Rocky, Die Hard, Lethal Weapon, Indiana Jones}.  Each
franchise comprises 4 movies; see Table~\ref{tab:actionmoviesstats} for summary statistics of the dataset. 

Each movie is decomposed into a list of shots, extracted with a shot 
boundary detector~\cite{massoudi2006video,potapov2014category}. Each shot is
tagged with zero, one or several labels corresponding to  the $11$ beat-categories 
(the label $\text{NULL}$ is assigned to shots with zero labels).
Note that the total footage for the dataset is 36.5 h, shorter than the total length in 
Table~\ref{tab:actionmoviesstats}. This is due to multiple labels. 
All categories are shown in
Figure~\ref{fig:categories}. 

Series of shots with the same category label are grouped together
in \emph{\beatevents{}} if they all depict the same scene (ie. same
characters, same location, same action, etc.). Temporally, we also allow a \beatevent{} to 
bridge gaps of a few unrelated shots. Beat-events belong to a single, non-NULL, beat-category.

The set of categories was inspired by the taxonomy of~\cite{stc}, and motivated
by the presence of common narrative structures and beats in action movies.
Indeed,
category definitions strongly rely on a split of the characters into ``good''
and ``bad'' tags, which is typical in such movies. 
Each category thus involves a fixed combination of
heroes and villains: both ``good'' and ``bad'' characters are present during
\classname{battle} and \classname {pursuit}, but only ``good'' heroes are 
present in the case of \classname{good argue good}.

Large intra-class variation is due to a number of factors: 
duration, intensity of action, objects and actors, \emph{and} different
scene locations, camera viewpoint, filming style.  For ambiguous cases we used
the ``difficult'' tag.

\subsection{Annotation protocol}

The annotation process was carried out in two passes by
  three researchers.  Ambiguous cases were discussed
  and resulted in a clear annotation protocol. 
In the first pass we manually annotated each shot with zero, one or several 
of the $11$ beat-category labels. 
In the second one we annotated the \beatevents{} by specifying
their category, beginning and ending shots. 
We tolerated gaps of 1-2 unrelated shots for
sufficiently consistent \beatevents{}. 
Indeed, movies are often edited into sequences 
of interleaved shots from two events, \eg between the main storyline and the ``B'' story. 

Some annotations are labeled as ``difficult'', if they are
semantically hard to detect, or ambiguous. For instance, in \franchise{Indiana Jones 3},
Indiana Jones engages in a romance with Dr. Elsa Schneider, who actually betrays him
to the ``bad guy''.  Romance between Indiana Jones and Dr. Elsa Schneider
is therefore ambiguous. We exclude these shots at training and evaluation time,
as in the Pascal evaluation protocol~\cite{Everingham10}.

Our beat-event annotations cover about 60~\% of the movie footage, which is much higher than
comparable datasets, see Table~\ref{tab:actionmoviesstats}.
This shows that the vocabulary we chose
is representative: the dataset is annotated densely.

\subsection{Highlighting structure of action movies}

Figure~\ref{fig:instances} shows the sequence of category-label annotations for several movies.
Some global trends are striking: \classname{victory good} occurs at the
end of movies; \classname{battle} is most prevalent in the last
quarter of movies; there is a pause in fast actions
(\classname{battle}, \classname{pursuit}) around the middle of the
movies. In movie script terms, this is the ``midpoint'' beat~\cite{stc}, 
where the hero is at a temporary high or low in the story.
In terms of \beatevent{} duration, \classname{joy bad} and
\classname{victory bad} are short, while \classname{pursuit} and
\classname{romance} are long. 
These trends can be learned by the
temporal re-scoring to improve the shot
classification results.

After careful analysis of the annotation, we find that 
\classname{battle}, \classname{despair good} and
\classname{pursuit} are the most prevalent beat-categories, with 4145, 3042 and
2416 instances respectively. Since it is a semantically high level
class, \classname{despair good} is most often annotated as difficult.
The co-occurrences of classes as annotations of the same shot follow
predictable trends: \classname{battle} co-occurs with
\classname{pursuit}, \classname{battle preparation},
\classname{victory good} and \classname{victory bad}. Interestingly
\classname{romance} is often found in combination with
\classname{despair good}. This is typical for movies of the ``Dude
with a problem'' type~\cite{stc}, where the hero must prove himself. 

Within each movie franchise, a shared structure may appear. For instance, 
in \franchise{Rocky}, the \classname{battle preparation} occurs in 
the last quarter of the movie, and there is no \classname{pursuit}.

\subsection{Evaluation protocol}

In the following, we propose two types of train/test splits and two performance measures
for our Action Movie Franchises dataset.

\paragraph{Data splits.}

We consider two different types of splits over the 20 movies; see
Figure~\ref{fig:splits}. They both come in 5 folds of 16 training
movies and 4 test movies. All movies appear once as a test movie.
In the ``leave one franchise out'' setting,
all movies from a single franchise are used as a test set. In ``leave
4 movies out'', a single movie from each franchise is used as test.
This allows to evaluate if our classifiers are specific to a franchise or 
generalize well across franchises. 

\newcommand{\cb}[1]{{%
\setlength{\fboxsep}{0pt}%
\setlength{\fboxrule}{1pt}%
\fbox{\textcolor{#1}{\rule{3mm}{3mm}}}%
}}

\newcommand{\cbx}[2]{\scalebox{0.9}{%
\setlength{\fboxsep}{0pt}%
\setlength{\fboxrule}{1pt}%
\fbox{\colorbox{#1}{\rule[-0.1em]{0pt}{0.8em}\,#2\,}}%
}}

\begin{figure}
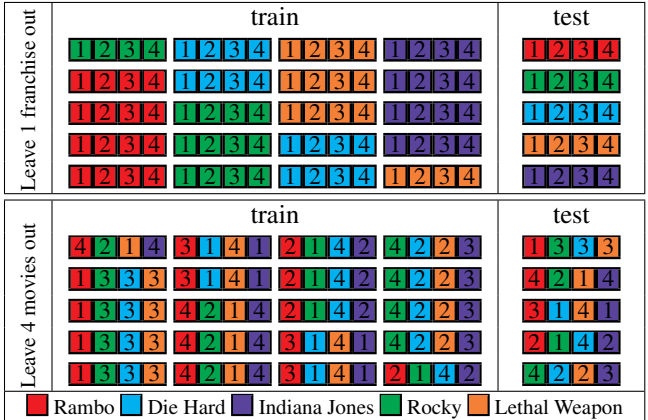
\begin{tabular}{|c|c|c|}
\hline
\raisebox{-2.2cm}[0pt][0pt]{\rotatebox{90}{\scalebox{0.8}{{Leave 1 franchise out}}}}
&train & test \\
&
\cbx{Green}{1}\cbx{Green}{2}\cbx{Green}{3}\cbx{Green}{4}
\cbx{ProcessBlue}{1}\cbx{ProcessBlue}{2}\cbx{ProcessBlue}{3}\cbx{ProcessBlue}{4}
\cbx{Orange}{1}\cbx{Orange}{2}\cbx{Orange}{3}\cbx{Orange}{4}
\cbx{Violet}{1}\cbx{Violet}{2}\cbx{Violet}{3}\cbx{Violet}{4}
&
\cbx{Red}{1}\cbx{Red}{2}\cbx{Red}{3}\cbx{Red}{4}
\\
&
\cbx{Red}{1}\cbx{Red}{2}\cbx{Red}{3}\cbx{Red}{4}
\cbx{ProcessBlue}{1}\cbx{ProcessBlue}{2}\cbx{ProcessBlue}{3}\cbx{ProcessBlue}{4}
\cbx{Orange}{1}\cbx{Orange}{2}\cbx{Orange}{3}\cbx{Orange}{4}
\cbx{Violet}{1}\cbx{Violet}{2}\cbx{Violet}{3}\cbx{Violet}{4}
&
\cbx{Green}{1}\cbx{Green}{2}\cbx{Green}{3}\cbx{Green}{4}
\\
&
\cbx{Red}{1}\cbx{Red}{2}\cbx{Red}{3}\cbx{Red}{4}
\cbx{Green}{1}\cbx{Green}{2}\cbx{Green}{3}\cbx{Green}{4}
\cbx{Orange}{1}\cbx{Orange}{2}\cbx{Orange}{3}\cbx{Orange}{4}
\cbx{Violet}{1}\cbx{Violet}{2}\cbx{Violet}{3}\cbx{Violet}{4}
&
\cbx{ProcessBlue}{1}\cbx{ProcessBlue}{2}\cbx{ProcessBlue}{3}\cbx{ProcessBlue}{4}
\\
&
\cbx{Red}{1}\cbx{Red}{2}\cbx{Red}{3}\cbx{Red}{4}
\cbx{Green}{1}\cbx{Green}{2}\cbx{Green}{3}\cbx{Green}{4}
\cbx{ProcessBlue}{1}\cbx{ProcessBlue}{2}\cbx{ProcessBlue}{3}\cbx{ProcessBlue}{4}
\cbx{Violet}{1}\cbx{Violet}{2}\cbx{Violet}{3}\cbx{Violet}{4}
&
\cbx{Orange}{1}\cbx{Orange}{2}\cbx{Orange}{3}\cbx{Orange}{4}
\\
&
\cbx{Red}{1}\cbx{Red}{2}\cbx{Red}{3}\cbx{Red}{4}
\cbx{Green}{1}\cbx{Green}{2}\cbx{Green}{3}\cbx{Green}{4}
\cbx{ProcessBlue}{1}\cbx{ProcessBlue}{2}\cbx{ProcessBlue}{3}\cbx{ProcessBlue}{4}
\cbx{Orange}{1}\cbx{Orange}{2}\cbx{Orange}{3}\cbx{Orange}{4}
&
\cbx{Violet}{1}\cbx{Violet}{2}\cbx{Violet}{3}\cbx{Violet}{4}
\\

\hline
\hline
\raisebox{-2.2cm}[0pt][0pt]{\rotatebox{90}{\scalebox{0.8}{{Leave 4 movies out}}}}
&train & test \\

&
\cbx{Red}{4}\cbx{Green}{2}\cbx{Orange}{1}\cbx{Violet}{4}
\cbx{Red}{3}\cbx{ProcessBlue}{1}\cbx{Orange}{4}\cbx{Violet}{1}
\cbx{Red}{2}\cbx{Green}{1}\cbx{ProcessBlue}{4}\cbx{Violet}{2}
\cbx{Green}{4}\cbx{ProcessBlue}{2}\cbx{Orange}{2}\cbx{Violet}{3}
&
\cbx{Red}{1}\cbx{Green}{3}\cbx{ProcessBlue}{3}\cbx{Orange}{3}
\\
&
\cbx{Red}{1}\cbx{Green}{3}\cbx{ProcessBlue}{3}\cbx{Orange}{3}
\cbx{Red}{3}\cbx{ProcessBlue}{1}\cbx{Orange}{4}\cbx{Violet}{1}
\cbx{Red}{2}\cbx{Green}{1}\cbx{ProcessBlue}{4}\cbx{Violet}{2}
\cbx{Green}{4}\cbx{ProcessBlue}{2}\cbx{Orange}{2}\cbx{Violet}{3}
&
\cbx{Red}{4}\cbx{Green}{2}\cbx{Orange}{1}\cbx{Violet}{4}
\\
&
\cbx{Red}{1}\cbx{Green}{3}\cbx{ProcessBlue}{3}\cbx{Orange}{3}
\cbx{Red}{4}\cbx{Green}{2}\cbx{Orange}{1}\cbx{Violet}{4}
\cbx{Red}{2}\cbx{Green}{1}\cbx{ProcessBlue}{4}\cbx{Violet}{2}
\cbx{Green}{4}\cbx{ProcessBlue}{2}\cbx{Orange}{2}\cbx{Violet}{3}
&
\cbx{Red}{3}\cbx{ProcessBlue}{1}\cbx{Orange}{4}\cbx{Violet}{1}
\\
&
\cbx{Red}{1}\cbx{Green}{3}\cbx{ProcessBlue}{3}\cbx{Orange}{3}
\cbx{Red}{4}\cbx{Green}{2}\cbx{Orange}{1}\cbx{Violet}{4}
\cbx{Red}{3}\cbx{ProcessBlue}{1}\cbx{Orange}{4}\cbx{Violet}{1}
\cbx{Green}{4}\cbx{ProcessBlue}{2}\cbx{Orange}{2}\cbx{Violet}{3}
&
\cbx{Red}{2}\cbx{Green}{1}\cbx{ProcessBlue}{4}\cbx{Violet}{2}
\\
&
\cbx{Red}{1}\cbx{Green}{3}\cbx{ProcessBlue}{3}\cbx{Orange}{3}
\cbx{Red}{4}\cbx{Green}{2}\cbx{Orange}{1}\cbx{Violet}{4}
\cbx{Red}{3}\cbx{ProcessBlue}{1}\cbx{Orange}{4}\cbx{Violet}{1}
\cbx{Red}{2}\cbx{Green}{1}\cbx{ProcessBlue}{4}\cbx{Violet}{2}
&
\cbx{Green}{4}\cbx{ProcessBlue}{2}\cbx{Orange}{2}\cbx{Violet}{3}
\\

\hline
\multicolumn{3}{|c|}{
\scalebox{0.8}{\makebox{
\cb{Red} Rambo 
\cb{ProcessBlue} Die Hard 
\cb{Violet} Indiana Jones 
\cb{Green} Rocky 
\cb{Orange} Lethal Weapon
}}
}\\
\hline
\end{tabular}
\caption{The two  types of split for evaluation. In addition to the train/test splits, the training videos are also split in 4 \emph{sub-folds}, that are used for cross-validation and CRF training purposes.}
\label{fig:splits}
\end{figure}

\paragraph{Classification setting.}
In the classification setting, we evaluate the accuracy of beat-category prediction 
at the shot level. Since a shot can have several labels, we
adopt the following evaluation procedure. For a given shot
with $n>0$ ground-truth labels (in general $n=1$, but the number of labels can be up to $4$),
we retain the best $n$ predicted beat-categories (out of 11, according to their
confidence scores).  
Accuracy is then measured independently for each beat-category as the 
proportion of ground-truth shots which are correctly labeled.
We finally average accuracies over all categories, 
and report the mean and the standard deviation over the 5 cross-validation
splits.

\paragraph{Localization setting.}
In the localization setting, we evaluate the temporal agreement between 
ground-truth and predicted \beatevents{} for each beat-category. 
A detection, consisting of a temporal segment, a category label and a confidence score,
is tagged positive if there exists a ground-truth \beatevent{} with 
an intersection-over-union score~\cite{Everingham10} over 0.2. 
If the ground-truth \beatevent{} is tagged as ``difficult'' it does not count as positive nor negative. 
The  performance is measured for each beat-category 
in terms of average precision (AP) over all \beatevents{} in the test fold, 
and the different APs are averaged to a mAP measure.


\section{Shot and beat-event classification}

The proposed approach consists of 4 stages.
First, we compute high-dimensional shot descriptors
for different visual and audio modalities, called \emph{channels}. Then, we learn linear SVM
classifiers for each channel. At the late fusion stage, 
we take the linear combination of the channel scores.
Finally, predictions are refined by leveraging the temporal
structure of the data and beat-events are localized. 

\subsection{Descriptors extraction}

For each shot from a movie, we extract different descriptors corresponding to different modalities.
For this purpose, we use a state-of-the art set of low-level descriptors~\cite{aly:hal-00904404,OVS13}.
It includes still image, face, motion and audio
descriptors:\\
{\bf Dense SIFT}~\cite{Low04} descriptors are extracted every 30'th frame. The
SIFTs of a frame are aggregated into a Fisher vector of 256 mixture
components, that is power- and L2-normalized~\cite{PSM10}. The shot
descriptor is the power- and L2 normalized average of the Fisher
descriptors from its frames. The output descriptor has 34559 dimensions.\\
{\bf Convolutional neural nets (CNN)} descriptors are extracted from every 30'th frame.
We run the image through a CNN~\cite{alexnips} trained on Imagenet 2012,
using the activations from the first fully-connected layer as a description
vector (FC6 in 4096~dimensions). The implementation is based on DeCAF~\cite{donahue2013decaf} and
its off-the-shelf pre trained network.\\
{\bf Motion descriptors} are extracted for each shot. We extract improved dense
trajectory descriptors~\cite{hengwangICCV13}. The 4 components of the
descriptor (MBHx, MBHy, HoG, HoF) are aggregated into 4 Fisher vectors that
are concatenated. This output is a 108544~D vector.\\
{\bf Audio descriptors} are based on MFCC~\cite{rabiner2007introduction} extracted for 25 ms
audio chunks with a step of 10 ms. They are enhanced by adding first and second
order temporal derivatives. The MFCCs are aggregated into a shot descriptor
using a Fisher aggregation, producing a 20223 D vector. \\
{\bf Face descriptors} are obtained by first detecting faces in each frame using the Viola-Jones detector
from OpenCV~\cite{opencv_library}. Following the approach from~\cite{everingham2006buffy}, we join the 
detections into face tracks using the KLT tracker, allowing us to recover some missed detections.
Each facial region is then described with a Fisher vector of dense SIFTs~\cite{Simonyan13} 
(16384 dimensions) which is power- and L2-normalized.
Finally, we average-pool all face descriptors within a shot and normalize again the result
to obtain the final shot descriptor.\\

Overall, each 2\,hr movie is processed in 6\,hr on a 16-core machine. We will make all
descriptors publicly available.

\subsection{Shot classification with SVMs}

\begin{figure}
\centering
\ig{0.8}{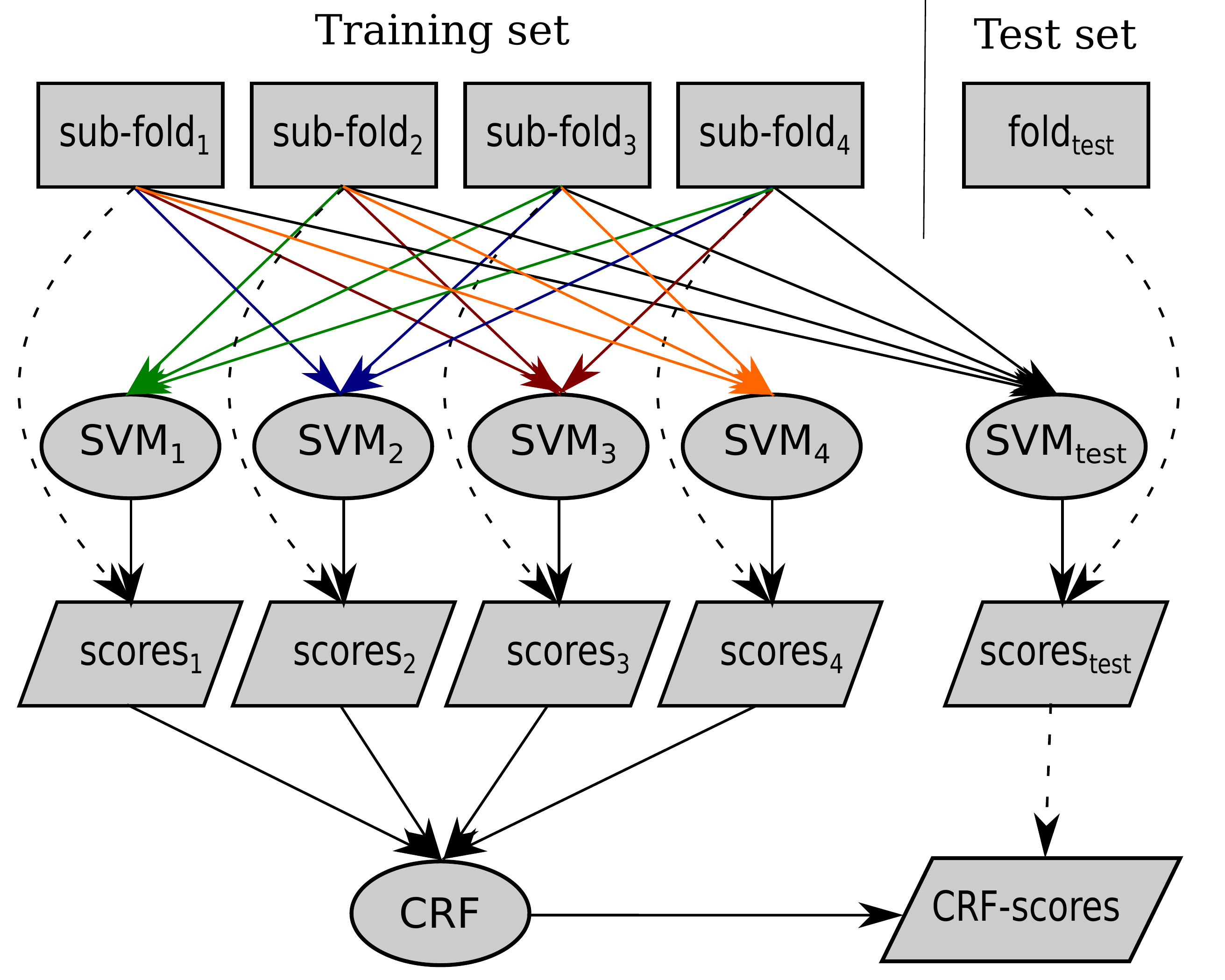}
\caption{Proposed training approach for one fold. In a first stage, SVMs $\text{SVM}_1...\text{SVM}_4$ are trained in leaving one sub-fold out 
of the training set, and are evaluated on the left-out sub-fold.  In a second stage, a CRF model is trained, taking the sub-fold SVMs scores as inputs. 
We then use all the training videos to train the final SVM model ($\text{SVM}_\textrm{test}$). 
The final model outputs scores on the test fold, which are then refined by the CRF model.
Note that each SVM training includes calibration using cross validation. }
\label{fig:hybrid}
\end{figure}

We now detail the time-blind detection method, that scores each shot
independently without leveraging temporal structure.

\vspace{-0.2cm}
\paragraph{Per-channel training of SVMs.}
The 5 descriptor channels are input separately to the SVM training. For each
channel and for each beat-category, we use all shots annotated as non-difficult as positive examples
and all other shots (excluding difficult ones) as negatives
to train a shot classifier. 
We use a linear SVM and cross-validate the $C$ parameter, independently for
each channel. We compute one classifier $\text{SVM}_\textrm{test}$ per fold, and 4 additional classifiers $\text{SVM}_1...\text{SVM}_4$ corresponding to sub-folds,  
see Figure~\ref{fig:hybrid}.

\vspace{-0.2cm}
\paragraph{Late fusion of per-channel scores}
The per-channel scores are combined linearly into a shot score. For one
fold, the linear combination coefficients are estimated using the sub-fold
scores. We use a random search over the 5D space of coefficients to find the one
that maximizes the average precision over the sub-folds. This optimization is
performed jointly over all classes (shared weights), which was found to be
better to reduce the variability of the weights. 

\subsection{Leveraging temporal structure}

We leverage the temporal structure to improve the
performance of the time-blind detection/localization method, using a
conditional random field (CRF)~\cite{lafferty2001conditional}. We consider a
CRF that takes the SVM scores as inputs. 
The CRF relies on a linear chain model. Unary potentials correspond
to votes for the shot labels, while binary potentials model the probability of
the sequences. 

We model a video with a linear chain CRF. 
It consists of latent nodes
$y_i \in \mathcal{Y}, i=1,\ldots,n$
that correspond to shot labels. Similar to HMM, each node $y_i$ has a
corresponding input data point $x_i \in \mathbb{R}^d$. 
Variables $x_i$ are always observed, whereas $y_i$
are known only for training data. 
An input data point $x_i \in \mathbb{R}^d$
corresponds to the shot descriptor, which 
in our case is the 11-D vector of L2-normalized SVM scores for each beat-category. 
The goal is to infer probabilities of shot
labels for the test video.

The CRF model for one video is defined as:
\begin{multline*}
    \log p(Y | X; \bm{\lambda}, \bm{\mu}) = 
    \sum_{i=1}^n \bm{\lambda}^T \bm{f}(y_i, X) + 
    \sum_{i=1}^{n-1} \bm{\mu}^T \bm{g}(y_i,y_{i+1}, X),
\end{multline*}
where the inputs are $X = \{x_1,\ldots,x_n\}$ 
and the outputs $Y = \{y_1,\ldots,y_n\}$. 
We use the
following feature (in the CRF literature sense) functions 
$f$ and $g$:
\begin{align*}
    f_k(y_i, X) &= x_{i,k} \delta(y_i, k) \\
    g_{k',k''}(y_i, y_{i+1}, X) &= \delta(y_i, k') \delta(y_{i+1}, k'') 
\end{align*}
where $x_{i,k}$ is the classification score of shot $i$ for category $k$,
$\delta(x,y)$ is 1 when $x=y$ and 0 otherwise. Therefore, the log-likelihood becomes
\begin{align*}
    \log p(Y  | X; \bm{\lambda}, \bm{\mu}) = 
        & \sum_{k \in \mathcal{Y}} \lambda_k \sum_{i=1}^n x_{i,k}\delta(y_i,k) + \\
        & \sum_{\substack{k',k''\in \mathcal{Y}\\(k',k'')\ne(c,c)}}
            \mu_{k',k''} \sum_{i=1}^{n-1} \delta(y_i,k') \delta(y_{i+1},k'')
\end{align*}
We take $x_i$ from SVM classifiers trained using cross
validation on the training data.  
The CRF is learned by minimizing the negative
log-likelihood in order to estimate $\bm{\lambda}$ and $\bm{\mu}$.

At test time, the CRF inference outputs marginal conditional probabilities
$p(y_i | X), i=1,\ldots,n$.

\newcommand{\std}[1]{{\scriptsize \it $\pm$ #1}}
\newcommand{\colhead}[1]{\rotatebox{90}{\begin{minipage}{2cm}\flushleft #1\end{minipage}}}
\begin{table*}
\centering
\scalebox{0.85}{
\begin{tabular}{|l|rrrrrrrrrrrr|}
\hline
&
\colhead{pursuit} & 
\colhead{battle} & 
\colhead{romance} & 
\colhead{victory good} & 
\colhead{victory bad} & 
\colhead{battle preparation} & 
\colhead{despair good} & 
\colhead{joy bad} & 
\colhead{good argue bad} & 
\colhead{good argue good} & 
\colhead{bad argue bad} &
 $\begin{array}{c}
\mbox{mean}\\
\mbox{accuracy}
\end{array}$
\\
\hline
  & \multicolumn{12}{c|}{\textbf{Leave 4 movies out}} \\

SIFT & 53.8 & 76.4 & 23.9 & 11.7 & 4.4 & 22.1 & 15.0 & 9.5 & 15.1 & 25.5 & 4.0 & 23.76 \std{\;\:5.26}  \\
CNN & {\bf 66.4} & 60.0 & 16.6 & 6.0 & 2.4 & 9.4 & 21.7 & 6.6 & 17.7 & 30.2 & 4.7 & 21.96 \std{\;\:5.91}  \\
dense trajectories &  58.5 & {\bf 85.2} & {\bf 38.0} & 12.7 & 6.2 & {\bf 28.0} & 19.5 & 11.6 & {\bf 18.8} & {\bf 40.4} & 1.8 & {\bf 29.15} \std{\;\:6.12}  \\
MFCC & 28.1 & 56.3 & 4.5 & {\bf 17.7} & {\bf 36.2} & 3.8 & {\bf 35.4} & {\bf 15.6} & 17.3 & 26.5 & 0.0 & 21.95 \std{13.97}  \\
Face descriptors & 47.9 & 58.1 & 8.6 & 12.7 & 11.4 & 17.3 & 9.3 & 3.2 & 6.2 & 22.3 & {\bf 4.7} & 18.35 \std{10.50}  \\
\hline
linear score combination & 63.9 & 89.2 & 32.3 & 14.0 & 11.4 & 18.6 & 26.0 & 12.1 & 18.0 & 44.3 & 1.8 & {\bf 30.15} \std{\;\:6.72}  \\
+ CRF & 76.0 & 91.2 & 57.6 & 19.9 & 1.0 & 41.4 & 43.1 & 9.6 & 25.1 & 44.8 & 0.0 & {\bf 37.25} \std{\;\:9.94}  \\
\hline 
& \multicolumn{12}{c|}{\textbf{Leave 1 franchise out}} \\
linear score combination & 57.8 & 83.6 & 13.0 & 14.9 & 9.6 & 3.8 & 28.0 & 5.2 & 18.2 & 44.3 & 0.0 & 25.32 \std{\;\:7.40}  \\
+ CRF & 75.4 & 87.4 & 31.3 & 15.8 & 0.0 & 12.7 & 33.4 & 5.7 & 23.2 & 43.7 & 0.0 & 29.89 \std{12.11}  \\
\hline
\end{tabular}}
\caption{\label{tab:bigres}
Performance comparison (accuracy) for shot classification. Standard deviations are computed over folds. 
}
\end{table*}

\begin{table*}
\centering
\scalebox{0.85}{
\begin{tabular}{|l|rrrrrrrrrrrr|}
\hline
\textcolor{white}{linear score combination}  & \multicolumn{12}{c|}{\textbf{Leave 4 movies out}} \\
CRF + thresholding & 34.6 & 38.9 & 22.6 & 14.6 & \;\:4.4 & 26.7 & \;\:6.4 & \;\:4.6 & 12.2 & 16.9 & \;\:0.6 & 16.59 \std{\;\:6.82} \\
\hline 
\hline
& \multicolumn{12}{c|}{\textbf{Leave 1 franchise out}} \\
CRF + thresholding & 36.8 & 36.5 & 28.9 & 14.3 & 4.5 & 1.7 & 4.2 & 5.2 & 6.5 & 13.5 & 3.7 & 14.16 \std{\;\:6.84} \\
\hline
\end{tabular}}
\caption{\label{tab:bigresloc}
Performance comparison (average precision) for \beatevent{} localization. 
}
\end{table*}


\begin{figure}
    \ig{1}{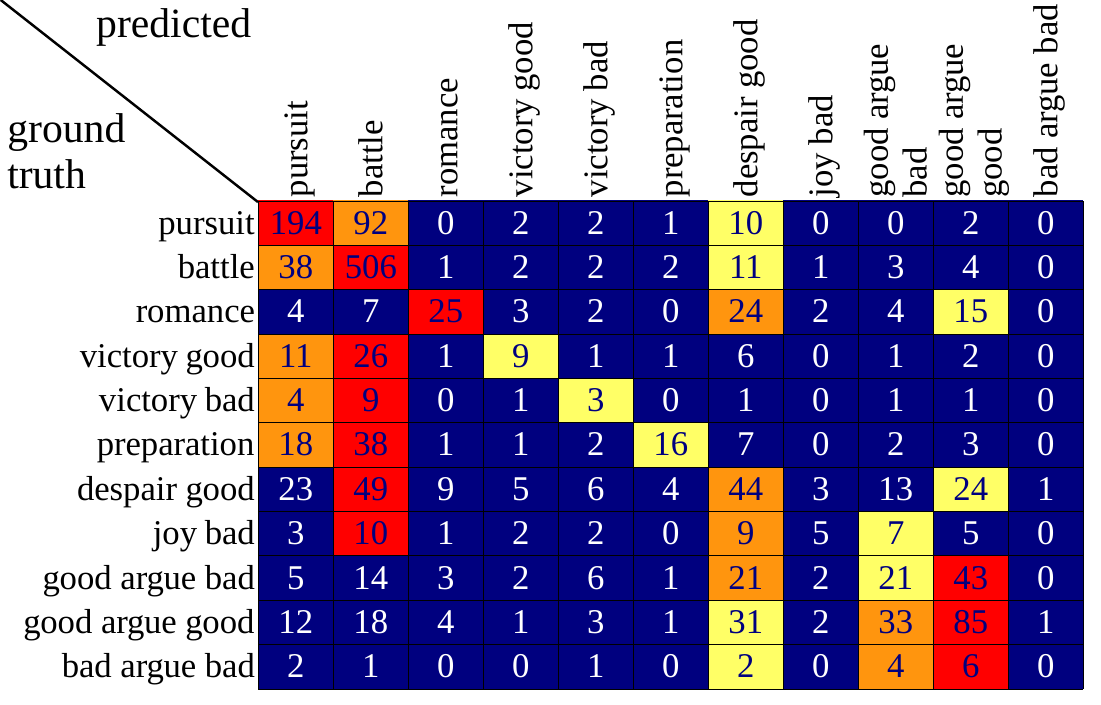}
    \caption{Confusion matrix for shot classification with SVM and linear score combination
for the ``leave 4 movies out'' setting. 
}
    \label{fig:confmat} 
\end{figure}

\subsection{Beat-event localization}

The final step consists in localizing instances of a \beatevent{} in a movie, 
given confidence scores output by the CRF. 
To that aim, shots must be grouped into segments, 
and a score must be assigned to the segments. We 
create segments by joining consecutive shots for which CRF confidence is above 30\%  
of its maximum over the movie. 
The segment's score is the average of these shot confidences.

\begin{figure*}[t!]
\newcommand{\myhline}[0]{\rule{\linewidth}{0.4pt}}
\centering
\includegraphics[width=\linewidth]{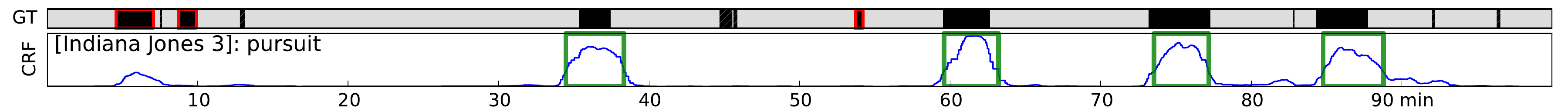} 
\includegraphics[width=\linewidth]{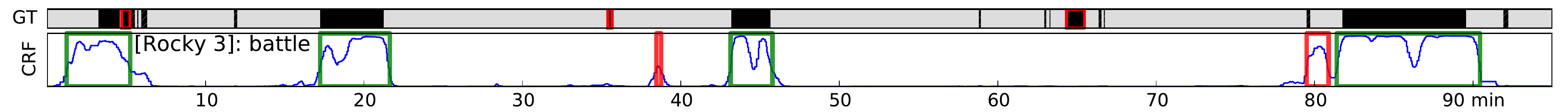} 
\includegraphics[width=\linewidth]{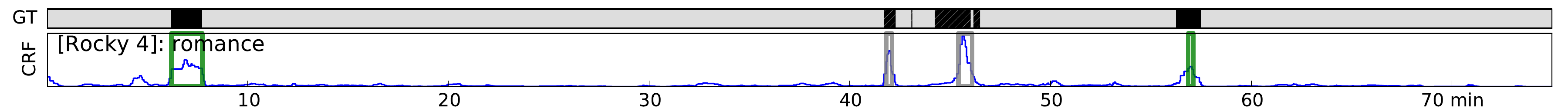} 
\includegraphics[width=\linewidth]{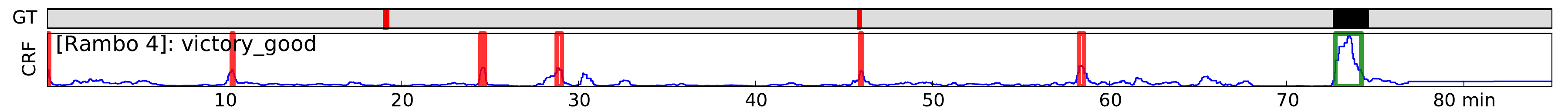} 
\includegraphics[width=\linewidth]{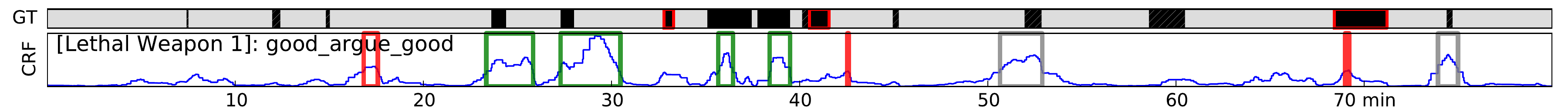}
\caption{Example of localization results, for several beat-categories and movies. 
For each plot, detected \beatevents{} are indicated with bold rectangles (green/gray/red indicate correct/ignored/wrong detections).
Ground-truth (GT) annotations are indicated above (\beatevents{} marked as difficult appear hatched), and 
likewise missed detections are highlighted in red.
Most often, occurrences of the \beatevents{} are rather straightforward to localize given the CRF scores. }
\label{fig:temporalscores}
\end{figure*}

Note that the CRF produces smoother scores over time for events that
occur at a slower rhythm, see Figure~\ref{fig:temporalscores}. For
example ``good argue good'' lasts usually longer than ``joy bad'', because the
villain is delighted for a short time only. The CRF smoothing
modulates the length of estimated segments: smoother curves
produce longer segments, as expected.


\section{Experiments}
\label{sec:exps}

After validating the processing chain on a standard dataset, we report
classification and localization performance.

\subsection{Validation of the classification method}

To make sure that our descriptors and classification chain is
reliable, we run it on the small Coffee \& Cigarettes~\cite{cac}
dataset, and compare the results to the state-of-the-art method
of Oneata et al.~\cite{OVS13}. For this experiment, we score fixed-size segments
and use their non-maximum suppression method NMS-RS-0. We
obtain 65.5~\% mAP for the ``drinking'' action and 45.4~\% mAP for 
``smoking'', which is close to their performance (63.9~\% and 50.5~\% respectively).

\subsection{Shot classification}

Table~\ref{tab:bigres} shows the classification performance at the shot-level 
on the two types of splits. 
The low-level descriptors that are most useful in this context are the
dense trajectories descriptors. Compared to setups like
Trecvid MED or Thumos~\cite{2014trecvidover,THUMOS14}, the relative
performance of audio descriptors (MFCC) is high, overall the same as for \eg CNN.
This is because Hollywood action movies have well controlled
soundtracks that almost continuously plays music: the rhythm and tone
of the music indicates the theme of the action occurring on screen.
Therefore, the MFCC audio descriptors convey  high-level information
that is relatively easy to detect automatically. 

\begin{figure}
\centering
\ig{1}{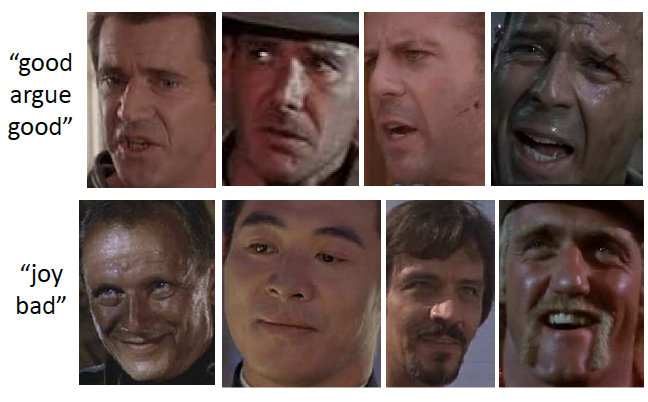}
\caption{Sample faces corresponding to shots for which the face classifier 
(\ie SVM trained on faces) scored much higher than the SIFT classifier (\ie trained on full images).
Similar facial expressions can be observed within each beat-category, which
suggests that our face classifier learns to recognize human expressions to some extent. }
\label{fig:faces}
\end{figure}

The face descriptor can be seen as a variant of SIFT, restricted to
facial regions. 
The face channel classifier outperforms SIFT in three categories. Upon inspection,
we noticed however that only a fraction of shots actually contain 
exploitable faces (\eg frontal, non-blurred, unoccluded and large enough),
which may explain the lower performance for other categories. 
The performance of the face channel classifier may be attributed
to a rudimentary facial expression recognition property: the faces of heroes arguing
with other good characters can be distinguished from the grin of the villain
in \classname{joy bad}; see Figure~\ref{fig:faces}. 

The 4 least ambiguous beat-categories (\classname{pursuit},
\classname{battle}, \classname{battle preparation} and
\classname{romance}) are detected most reliably. They account for more
than half of the annotated shots. The other categories are typically
interactions between people, which are defined 
by identity and speech rather than motion or music. 
The confusion matrix  in Figure~\ref{fig:confmat} shows that
verbal interactions like ``good argue good'' and ``good argue bad'' are often confused.

The ``leave-4-movies out'' setting obtains significantly better
results than ``Leave-1-franchise out'', meaning that having seen
movies from a franchise makes it easier to recognize what is happening
in a new movie of the franchise:
Rambo does not fight in the same way as Rocky.
Finally, the CRF allows to leverage temporal structure using the temporally
dense annotations, improving the classification performance by 7~points.

\vspace{-0.1cm}
\subsection{Beat-event localization}

\vspace{-0.1cm}
Table~\ref{tab:bigresloc} gives results for beat-event localization. 
We observe that the performance is low for the least frequent 
actions. Indeed, for 8 out of 11 categories, the performance is below 
15\% AP. Per-channel results are not provided due to lack of space, but
their relative performance is similar to the classification ones.
Figure~\ref{fig:temporalscores} displays localization results for different
beat-categories. Categories, such as battle and pursuit, 
are localized reliably. Semantic categories, such as romance, victory good and good
argue good are harder to detect.
More advanced features could improve the results for these 
events. 
Indeed, recognition of characters, their pose and speech appear necessary.


\vspace{-0.1cm}
\section{Conclusion}

\vspace{-0.1cm}
Despite the explosion of user-generated video content, people are
still watching professionally produced videos most of the time.
Therefore, the analysis of this kind of footage will remain an
important task. 
In this context, Action Movie Franchises appears as a challenging benchmark. 
The
annotated classes range from reasonably easy to recognize
(\classname{battle}) to very difficult and semantic (\classname{good
  argue bad}). 
We also provide baseline results from a method that builds on state-of-the-art descriptors and classifiers.
Therefore, we expect it to be a valuable test case in
the coming years. 
We will provide the complete annotations and evaluation scrips upon
publication.

{\small
\bibliographystyle{ieee}
\bibliography{refs}
}

\end{document}